\documentclass{article}
\pdfoutput=1
\usepackage{microtype}
\usepackage{graphicx}
\usepackage{subfigure}
\usepackage{booktabs} % for professional tables

\usepackage{amsmath,amsfonts,bm}

\def\eqref#1{equation~\ref{#1}}
\def\1{\bm{1}}

\DeclareMathAlphabet{\mathsfit}{\encodingdefault}{\sfdefault}{m}{sl}
\SetMathAlphabet{\mathsfit}{bold}{\encodingdefault}{\sfdefault}{bx}{n}

\newcommand{\fig}[1]{Figure~\ref{#1}}

\def\model{RoleNRules\xspace}
\def\modelshort{RNR\xspace}

\usepackage{mfirstuc,tabulary}

\usepackage{color,xcolor}
\usepackage{epsfig}
\usepackage{graphicx}

\usepackage{adjustbox}
\usepackage{array}
\usepackage{booktabs}
\usepackage{colortbl}
\usepackage{float,wrapfig}
\usepackage{hhline}
\usepackage{multirow}
\usepackage[toc,page]{appendix}

\usepackage{amsmath,amsfonts,amsthm,amssymb}
\usepackage{bm}
\usepackage{nicefrac}
\usepackage{microtype}

\usepackage{changepage}
\usepackage{extramarks}
\usepackage{fancyhdr}
\makeatletter \def\@fancyvbox#1#2{\vbox{#2}} \makeatother
\usepackage{lastpage}
\usepackage{setspace}
\usepackage{soul}
\usepackage{xspace}
\usepackage{indentfirst}

\usepackage{pifont}

\usepackage[breaklinks=false]{hyperref}

\usepackage{algorithm, algorithmic}
\usepackage{listings}
\usepackage{enumerate}
\usepackage{mathtools}
\usepackage{pdfpages}
\usepackage[symbol]{footmisc}

\usepackage{etoolbox}
\makeatletter
\AfterEndEnvironment{algorithm}{\let\@algcomment\relax}
\AtEndEnvironment{algorithm}{\kern2pt\hrule\relax\vskip3pt\@algcomment}
\let\@algcomment\relax
\newcommand\algcomment[1]{\def\@algcomment{\footnotesize#1}}
\renewcommand\fs@ruled{\def\@fs@cfont{\bfseries}\let\@fs@capt\floatc@ruled
  \def\@fs@pre{\hrule height.8pt depth0pt \kern2pt}%
  \def\@fs@post{}%
  \def\@fs@mid{\kern2pt\hrule\kern2pt}%
  \let\@fs@iftopcapt\iftrue}
\makeatother
\usepackage{arydshln}

\usepackage[accepted]{icml2024}

\usepackage{amsmath}
\usepackage{amssymb}
\usepackage{mathtools}
\usepackage{amsthm}

\usepackage[capitalize,noabbrev]{cleveref}

\theoremstyle{plain}

\theoremstyle{definition}

\theoremstyle{remark}

\usepackage[textsize=tiny]{todonotes}

\icmltitlerunning{RNR: Teaching Large Language Models to Follow Roles and Rules}

\begin{document}

\twocolumn[
\icmltitle{
RNR: Teaching Large Language Models to Follow Roles and Rules
}
\icmlsetsymbol{equal}{*}

\begin{icmlauthorlist}
\icmlauthor{Kuan Wang}{equal,sch}
\icmlauthor{Alexander Bukharin}{equal,sch}
\icmlauthor{Haoming Jiang}{comp}
\icmlauthor{Qingyu Yin}{comp}
\icmlauthor{Zhengyang Wang}{comp}
\icmlauthor{Tuo Zhao}{sch}
\icmlauthor{Jingbo Shang}{comp}
\icmlauthor{Chao Zhang}{comp}
\icmlauthor{Bing Yin}{comp}
\icmlauthor{Xian Li}{comp}
\icmlauthor{Jianshu Chen}{comp}
\icmlauthor{Shiyang Li}{comp}
\end{icmlauthorlist}


\icmlaffiliation{sch}{Georgia Institute of Technology}
\icmlaffiliation{comp}{Amazon}
\icmlcorrespondingauthor{Shiyang Li}{syangli@amazon.com}

\icmlkeywords{Machine Learning, ICML}

\vskip 0.3in
]



\printAffiliationsAndNotice{\icmlEqualContribution and work was done during Kuan and Alexander' internship at Amazon.} % otherwise use the standard text.

\begin{abstract}

  Instruction fine-tuning (IFT) elicits instruction following capabilities and steers the behavior of large language models (LLMs) via supervised learning. However, existing models trained on open-source IFT datasets only have the ability to follow instructions from users, and often fail to follow complex role and rules specified by developers, a.k.a. system prompts. The ability to follow these roles and rules is essential for deployment, as it ensures that the model safely interacts with users within developer defined guidelines. To improve such role and rule following ability, we propose \model, an automated data generation pipeline that generates diverse roles and rules from existing IFT instructions, along with corresponding responses. This data can then be used to train models that follow complex system prompts. The models are evaluated on our newly created benchmarks for role and rule following ability, as well as standard instruction-following benchmarks and general NLP tasks. Our pipeline significantly improves role and rule following capability in LLMs, as evidenced by over 25\% increase in pass-rate on rule adherence, i.e. following all requirements, in our experiments with the Alpaca and Ultrachat datasets. Moreover, our models achieves this increase without any regression on popular instruction following benchmarks.

\end{abstract}

\section{Introduction}
\label{intro}

Instruction fine-tuning (IFT) ~\cite{mishra2021cross, ouyang2022training, muennighoff2022crosslingual, longpre2023flan, wang2023interactive, zhang2023chinese} has emerged as a powerful technique to elicit latent capabilities and steer the behaviors of Large Language Models (LLMs)~\cite{radford2019language, 2020t5, brown2020language, wei2022emergent, scao2022bloom, qin2023chatgpt, openai2023gpt4, chowdhery2022palm}. By fine-tuning with carefully curated instruction-response pairs, IFT enables LLMs to generate desirable responses to user instructions \cite{bach2022promptsource, wang2022super, wang2022self, taori2023alpaca}. IFT has exploded in popularity due to its ease of use and low cost, allowing developers to build a wide variety of LLM-based products.

Current IFT research is mainly focused on handling user input instructions.
However, this overlooks the need to enable developers to control LLM behavior via specific role and rule requirements. This enhanced level of controllability is necessary, as developers often need to specify the LLM's role in their application (``\textit{You are an AI coding assistant}" or ``\textit{You are an AI tax advisor}") as well as specific rules (``\textit{Return the results in XML format}" or ``\textit{Respond in less than 300 words}"). To enable such customization ability, many closed-source LLMs like GPT-4
\footnote{\url{https://platform.openai.com/docs/guides/prompt-engineering/tactic-ask-the-model-to-adopt-a-persona}} 

and Claude 2.1 
\footnote{\url{https://docs.anthropic.com/claude/docs/how-to-use-system-prompts}} allow developers to input roles and rules as a system prompt, which will then be used to control the LLMs behavior when interacting with the user. However, the techniques used to enable such role and rule following ability are currently not available to the broader research community.

 Existing open-source IFT datasets commonly employ a fixed system prompt (e.g., \emph{``You are a helpful, respectful, and honest assistant.''}).
One workaround to enable role and rule following ability is to prepend these requirements to user inputs, but we find this approach is not very effective. This is because open source IFT models are not trained with such concatenated system-prompts and user inputs, and developer role and rule requirements are often complex and diverse. This approach therefore causes a significant shift between the training and evaluation. Therefore, we find that models often overlook or incorrectly interprets specific role and rule requirements.

\begin{figure*}[t]
% \vspace{15pt}
  \centering
  \includegraphics[width=0.99\textwidth]{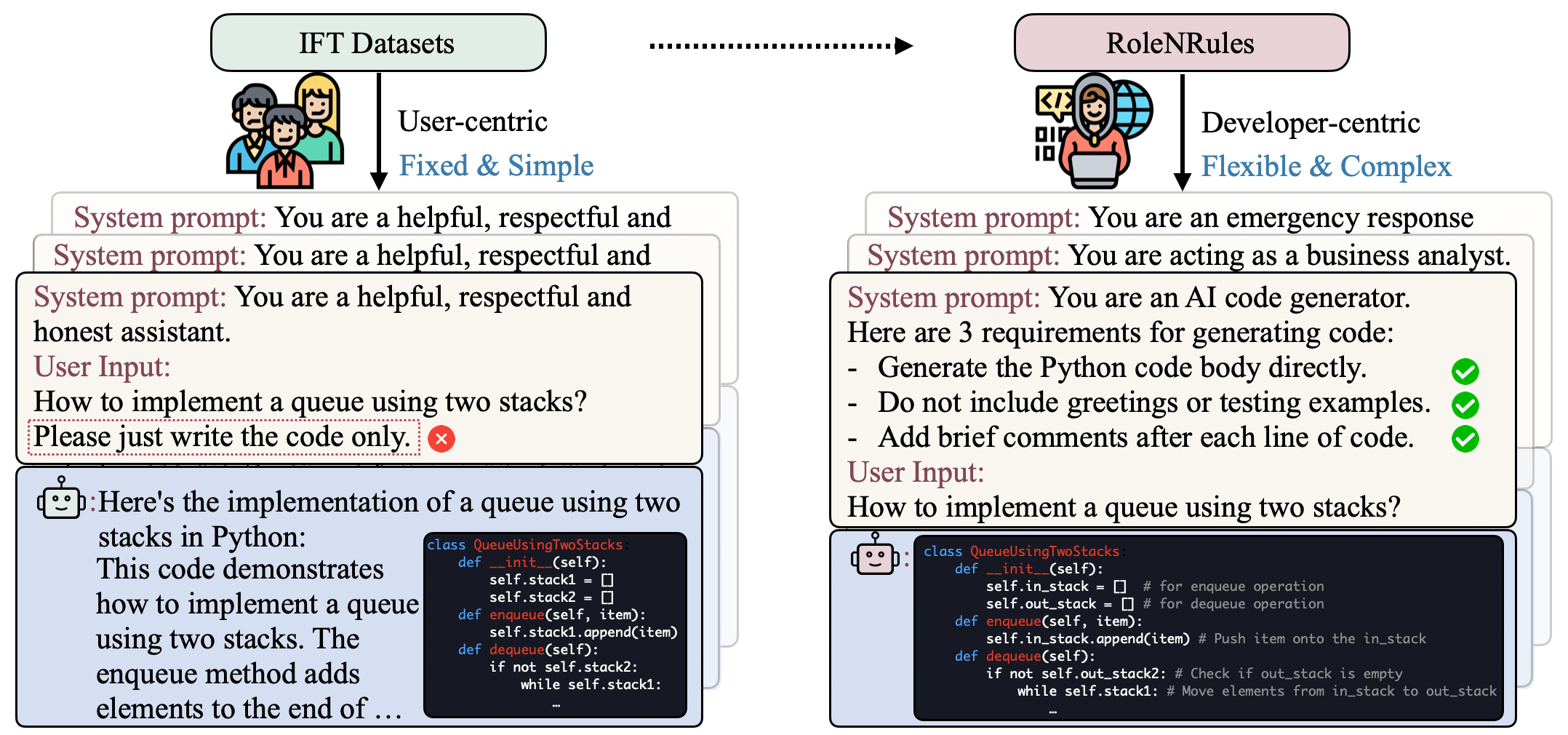}
  \caption{Generated response for a model trained with exisiting IFT datasets (left) and RNR (right). The model fine-tuned with existing IFT datasets ignore the user's extra requirement to write the code only. While the model trained with our \modelshort can follow all the requirements included in the complex system prompts. 
  }
    \label{fig:ift-challenge}
  
  % \label{fig:teaser}
  % \vspace{15pt}

\vspace{-15pt}
\end{figure*}

To illustrate this challenge, consider the example in Figure \ref{fig:ift-challenge}, where we ask a model trained on a popular open-source IFT dataset to respond to a system prompt and instruction. Even when the developer explicitly requires the system to act as a coder and generate code only, the system ignores this requirement and includes additional information. 
In contrast, closed-source models offer the ability to effectively adapt to various role and rule requirements. Therefore, there is a clear need for developing open-source IFT methods that support adaptable system prompts to enable developers' fine-grained model controls.

To bridge this gap, we propose \model (\modelshort), a scalable and automated data generation pipeline designed to enhance the ability of LLMs to follow developer instructions specified in the system prompt. More specifically, we use powerful language models to generate diverse and high quality (system-prompt, instruction, response) triplets from existing open-source IFT datasets. These triplets are then used for training the model, resulting in superior adaptability to system prompts. As demonstrated in Figure 
\ref{fig:ift-challenge}, the model trained with \modelshort can follow all the requirements defined inside the system prompts.

To generate diverse and high-quality (system prompt, instruction, response) triplets, the most challenging part is to create diverse and high-quality system prompts to avoid potential over-fitting of model training and hence improve model's adaptability to new roles and rules. A system prompt often has two parts:
1) \textbf{Role} -- which explicitly outlines the identity or persona that the LLM agent should assume during the interaction, e.g. ``\textit{You are a helpful coding assistant.}".
2) \textbf{Rules} -- which dictate the requirements, such as operational guidelines (e.g. ``\textit{Add brief comments after each line of code}") and objectives, which follow the role description. To generate diverse and high-quality system prompts, we prompt a powerful LLM to generate a novel role with a set of rules for each instruction in an open source IFT dataset. After that, we use the generated system prompt and original instruction to generate a response that follow the instruction while satisfying the system prompt. Through these steps, RNR generates a diverse dataset demonstrating role and rule following abilities, which can then be used to train system prompt following LLMs.

In addition to the \modelshort data generation pipeline, we also provide a benchmark specifically designed to assess the ability of LLMs to adhere to system prompts. This benchmark provides a set of challenging (system-prompt, instruction) pairs written by both humans and LLMs, as well as a technique to evaluate system-prompt following ability using LLM-based judges.

To validate the effectiveness of our \modelshort, we conduct extensive experiments using the Alpaca \citep{alpaca} and Ultrachat \citep{ding2023enhancing} datasets. We train Llama-2 \citep{touvron2023llama} models of various sizes on datasets generated by standard IFT techniques and by \modelshort. We find that \modelshort significantly boosts system-prompt following ability, improving the pass-rate by over 25\% compared to conventional instruction fine-tuning. In addition, models trained with \modelshort maintain the ability to follow complex instructions even when there is no specific system prompt, validating RNR's ability to train high performing and adaptive LLMs.

Our primary contributions are:

\begin{enumerate}
  \item \textbf{\model Generation Pipeline}: We design a general pipeline for automatic and scalable diverse role and rules IFT data generation. Our extensive experiments and human evaluations show that models trained on our data can enable model developers to have fine-grained control of model behaviors via system prompt engineering.
  
  \item \textbf{\model Benchmarks}: We design three high-quality benchmarks via human experts and LLMs to evaluate role and rule following capabilities of LLMs. We believe these datasets can further foster research on this topic.

  \item \textbf{\model Evaluation}: Further more, we design a LLM-based technique with a high level of human expert agreement to automatically evaluate the performance of models on the role and rules following benchmarks.

\end{enumerate}

\section{Method}
\label{sec:method}
We propose \model, a data generation pipeline for the automatic generation of \modelshort system prompts and the corresponding responses for IFT datasets. As illustrated in Figure \ref{fig:overview}, the \model pipeline augments an existing open-source IFT dataset into our \modelshort dataset with complex system-prompt data. Below, we describe the main processes in our pipeline.

\subsection{\model System Prompt}

Our method for improving role and rules following generates diverse and high quality system prompts from existing instruction tuning datasets. These system prompts are designed as a combination of a role description and a set of defined rules:

\begin{enumerate}
  \item \textbf{Role}:
        The role description explicitly specifies the identity or persona that the LLM should adopt during the interaction. For example, this role could be a technical assistant, customer service representative, data analyst, or any other persona with specific characteristics and knowledge.
        By defining the role, we guide the model to adopt a particular set of behaviors and language styles appropriate for that persona.
  \item \textbf{Rules}:
        Accompanying the role description are specific rules that set the boundaries and objectives of the interaction.
        These rules encompass operational guidelines, ethical considerations, data handling procedures, and specific goals to be achieved. For example, a rule might mandate prioritizing data privacy or adhering to industry-specific regulations.
\end{enumerate}

By integrating these two elements, we can simulate a wide variety of possible developer requirements.
In particular, the role descriptions make the LLM's response style customizable, while the rules ensure that the model operates within specific requirements. In addition, this approach enhances the LLM's ability to understand and execute complex and specialized tasks. And it also provides a pipeline for developers to tailor the model's responses for various professional scenarios.

\subsection{System Prompt Generation}
To generate unique system prompts comprised of different role and rules, we prompt powerful LLMs to generate role and rules based on an instruction from existing open-source datasets. We generate role and rules based on existing IFT datasets as existing IFT datasets contain instructions from a diverse range of subjects and tasks, which will in turn make our generated role and rules diverse.

To generate a role and rules based system prompt from an existing instruction, we sample a random variable $K$ from $0-8$ and ask Claude-2 to generate a role and $K$ rules based on a set of guidelines. This can be seen in Figure \ref{fig:overview}. These guidelines ensure that the role and rules are reasonable for a LLM to follow and that the role and rules are not overly specific. In addition, we provide three expert-generated demonstrations to help the LLM generate new roles and rules. Our complete prompt can be found in Appendix \ref{app:train_prompt}. In our experiments we use Claude-2 to generate the roles and rules, but in practice other powerful LLMs can be used.

\subsection{Response Generation}
Although existing IFT datasets contain responses to the instructions, likely, the instructions do not follow the generated role and rules. Therefore, once we generate a system prompt for a given instruction from the IFT dataset, we combine it with the instruction to form a new prompt for response generation. This is demonstrated in Figure \ref{fig:template}.
Notably, we only use the instructions from the IFT dataset without incorporating the original responses.
This approach prevents overfitting to the original responses and preserves diversity in the generated requirements.

\subsection{Fine-tuning}

\begin{figure}[ht]
% \vspace{-15pt}
  \vspace{-10pt}
  \centering
  \includegraphics[scale=0.34]{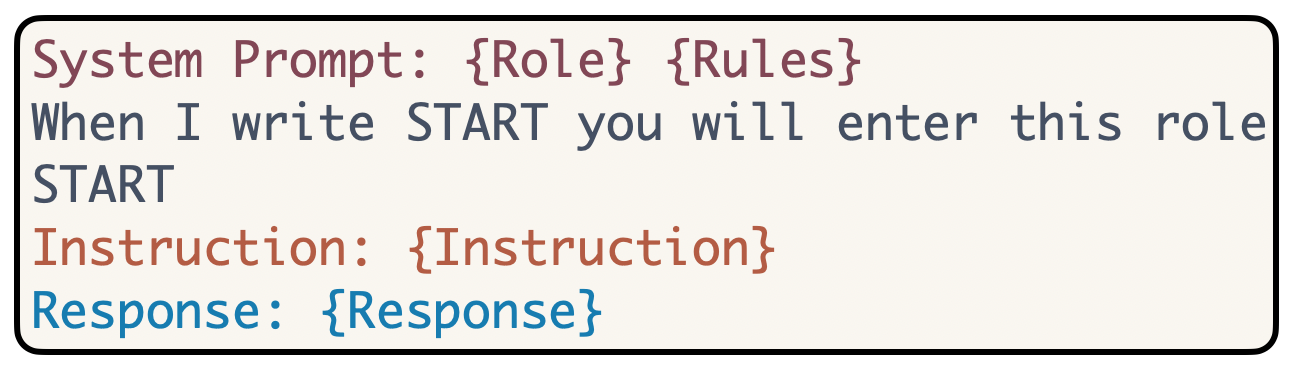}
  % \vspace{-10pt}
  \caption{We use this template to integrate the responses with the generated system prompts and instructions. }
  \label{fig:template}
\vspace{-10pt}
\end{figure}

 Once we have generated a dataset of (system prompt, instruction, response) triplets, we finetune a pre-trained LLM to maximize the probability of the response given the combined system prompt and instruction. To combine the system prompt and instruction, we follow the same template as in Figure \ref{fig:template}. This combined system prompt is used as the input for supervised fine tuning, where the target output is the response. \fig{fig:overview} is an overview of the \modelshort. In addition, we display examples of our training data in Appendix \ref{app:train-examples}.

\section{Experiments}

\subsection{Experimental Setup}

\begin{table*}[t]
% \vspace{-5pt}
\caption{Pass rate on the main evaluation sets. Llama-2-7B is the base model for RNR.}

\label{tab:main}
\centering
\scalebox{0.84}{

\begin{tabular}{| c | c c |c c |c c | c c |}
\hline RNR:Alpaca & \multicolumn{2}{| c |}{\textbf{RNR-Expert} } & \multicolumn{2}{ c |}{\textbf{RNR-Awesome} } & \multicolumn{2}{ c |}{\textbf{RNR-Alpacafarm} } & \multicolumn{2}{ c |}{\textbf{Average} }  \\ 
\cline{2-9}
 & Rule Level & Prompt Level & Rule Level & Prompt Level & Rule Level & Prompt Level & Rule Level & Prompt Level  \\
\hline

IFT: fixed-system & 53.62 & 39.53 & 58.39 & 28.6 & 72.57 & 43.21 & 61.53 & 37.11 \\
IFT: no-system & 64.76 & 57.78 & 67.59 & 39.05 & 74.41 & 44.47 & 68.92 & 47.10 \\
Rolenrule & \textbf{82.24} & \textbf{77.78} & \textbf{74.91} & \textbf{49.07} & \textbf{86.34} & \textbf{67.30} & \textbf{81.16} & \textbf{64.72} \\
\hline
\hline
 RNR:Ultrachat & Rule Level & Prompt Level & Rule Level & Prompt Level & Rule Level & Prompt Level & Rule Level & Prompt Level  \\
\hline

IFT: fixed-system & 61.18 & 48.89 & 57.51 & 29.10 & 76.14 & 49.56 & 64.94 & 42.52 \\
IFT: no-system & 80.50 & 70.21 & 63.07 & 36.49 & 75.23 & 51.09 & 72.93 & 52.60  \\
Rolenrule & \textbf{85.56} & \textbf{75.0} & \textbf{75.17} & \textbf{49.59} & \textbf{89.56} & \textbf{74.37} & \textbf{83.43} & \textbf{66.32} \\
\hline
\end{tabular}}
\vspace{-15pt}

\end{table*}

\textbf{Training Setup.} We present our main results on the Alpaca dataset \citep{taori2023alpaca} and the Ultrachat dataset \citep{ding2023enhancing}. For Alpaca, we generate (system prompt, instruction, response) triplets on the whole dataset and for Ultrachat, we sample a subset of size 50K to generate the triplets. For our main experiments, we use Llama 2 7B \citep{touvron2023llama} as our base model and fine-tune each model for two epochs with a batch size of 128 using a context length of 2048 and a learning rate of $2 \times 10^{-5}$.   

\textbf{Baselines.} To investigate the efficacy of RNR, we compare it with two IFT baselines. For both of these baselines we generate responses to the original training instructions with Claude-2, so that there is no discrepancy in the teacher model used for training. Once we generate these responses, we train models using (a) a fixed system prompt (``You are a helpful, harmless, and honest AI assistant") and (b) with no system prompt. These baselines are denoted as IFT: fixed-system and IFT: no-system, respectively.

\subsection{Benchmarks}
We design three benchmarks using the \modelshort format (system prompt + instruction)  to evaluate the effectiveness of \model:

\begin{enumerate}
  \item \textbf{RNR-Expert benchmark}: This is a small benchmark comprised of 50 data points with \modelshort system prompts and instructions handcrafted by domain experts, who are LLM researchers and are familiar with prompting techniques. The roles and rules in these system prompts cover diverse domains including programming, engineering, medicine, law, finance, movies, book recommendations, travel advice, and cooking recipes, etc. We remark that this dataset contains challenging and high quality examples.
  \item \textbf{RNR-Awesome benchmark}: This is benchmark has 495 data points based on the instruction collected by awesome-chatgpt-prompts~\cite{akin2023awesome}. For each instruction in awesome-chatgpt-prompts, we randomly select 6 samples from the RNR-Expert benchmark as the few-shot prompt, then leverage GPT-4 to generate the \model system prompt for this instruction. Given this system-prompt, we also use GPT-4 to generate 2 more possible instructions, so the amount of data points in RNR/Awesome is 3 times of the number of prompts in the awesome-chatgpt-prompts dataset.
  \item \textbf{RNR-AlpacaFarm benchmark}: This benchmark contain 805 datapoints, and the instructions are kept the same as the original AlpacaFarm \citep{dubois2023alpacafarm}. We also use 6 random samples from the RNR-Expert benchmark as the few-show prompt and leverage GPT-4 to generate the \model system prompt for each instruction in AlpacaFarm.
\end{enumerate}

\textbf{Evaluation.} To evaluate the system-prompt following ability of each model, we use large language models to calculate the number of rules and roles that are satisfied for each prompt. In particular, we develop a novel chain-of-thought \citep{wei2022chain} like prompt for evaluating \model following ability. More concretely, we ask the judge LLM to examine each role and rule requirement individually, and then aggregate these score into a final judgement. In addition, we include some few-shot examples to further boost performance. The exact prompt can be found in Appendix \ref{app:eval_prompt}. In our human study, we find that this evaluation strategy has a high level of agreements with an ensemble of human experts.

In our experiments, we use Claude-2.1 to calculate the percentage of rules in the system prompt that each model passes over each RNR benchmark dataset. We remark that we use Claude-2.1 as the evaluator instead of Claude-2 to mitigate the effect of LLM judges favoring their own output. We calculate the rule-level pass percentage (defined as the percent of rules each model passes) and the prompt-level pass percentage (defined as the percent of prompts where all rules are satisfied). 

Beyond measuring system prompt following ability, we also measure the ability of each model to follow user-inputted rules with IFEval \citep{zhou2023instruction}, role-playing ability with RoleBench \citep{wang2023rolellm}, instruction following ability with Alpacafarm \citep{dubois2023alpacafarm}, and reasoning ability with benchmarks such as MMLU \citep{hendrycks2020measuring}, BBH \citep{suzgun2022challenging}, and DROP \citep{dua2019drop}. For IFEval, we can programmatically evaluate the percent of rules each model passes. For RoleBench and Alpacafarm we compare outputs from the RNR model to outputs from the baselines to evaluate which model better follows the defined role or better follows the instruction. To judge each pair of responses, we use Claude-2.1 to output its preferred response in a similar manner to \citep{dubois2023alpacafarm}. In order to mitigate the positional bias of the judge LLM, we show it each comparison in both possible orders (with RNR first and with RNR second). We only evaluate our Alpaca based models on RoleBench and Alpacafarm to reduce evaluation costs. 

\subsection{Main Results}
Our main results on the RNR benchmark can be found in Table \ref{tab:main}. The results show that \modelshort can significantly improve system-prompt following ability compared to the two IFT baselines for both the Alpaca and Ultrachat datasets. In particular, we find that \modelshort improves prompt-level system-prompt following ability by 37.4\% on the Alpaca dataset and by 26.1\% for Ultrachat compared to the strongest baseline. These results indicate that \modelshort is an effective way to improve system-prompt following ability.

The results for IFEval can be found in Table \ref{tab:ifeval}. From Table \ref{tab:ifeval} we can see that RNR significantly boosts instruction following ability, even when the instruction is inputted by the user, indicating that improve system prompt following capability can positively transferred to user inputs.

For RoleBench, we find that RNR improves the winning score (defined as $\frac{\textrm{\# Win - \# Lose}}{\textrm{\#Win + \#Lose + \#Tie}} + 1$) by over 25\% compared to both IFT baselines on the Alpaca dataset. The results can be found in Table \ref{tab:rolebench}. This indicates RNR significantly enhances role-playing ability.

For evaluation on Alpacafarm, we do a head to head comparison of RNR:Alpaca and IFT:no-system. We find that the performance is similar for IFT:no-system and RNR, with RNR outperforming the IFT model $\sim$ 3.7\%. (See Appendix \ref{app:alpacafarm}). Therefore RNR can boost complex system-prompt following ability without hurting general instruction following performance. Finally, we demonstrate the performance of RNR on various benchmarks in Table \ref{tab:benchmark}. As expected, the benchmark performance is similar for all models, which shows that RNR retains general reasoning capabilities.

\begin{table}[htb]
\setlength{\tabcolsep}{1.2pt}

\caption{Pass rate on the IFEval. Llama-2-7B is the base model.}
% \vspace{5pt}
\label{tab:ifeval}
% \vspace{-15pt}
\centering
\scalebox{0.82}{
\begin{tabular}{| c | c c |c c |}
\hline & \multicolumn{2}{ c |}{\textbf{IFEval:Loose} } & \multicolumn{2}{ c |}{\textbf{IFEval:Strict} }  \\ 
\cline{2-5}
 Alpaca & {Rule Level} & Prompt Level & Rule Level & Prompt Level  \\
\hline

IFT: fixed-system & 39.69 & 27.73 & 32.25 & 19.69  \\
IFT: no-system & 39.69 & 26.80 & 32.01 & 19.41  \\
RNR &\textbf{51.68} & \textbf{39.57} & \textbf{43.65} & \textbf{31.61} \\
\hline
\hline
 Ultrachat & Rule Level & Prompt Level & Rule Level & Prompt Level  \\
\cline{1-5}
IFT: fixed-system & 39.69 & 27.34 & 30.70 & 19.04  \\
IFT: no-system & 40.17 & 28.10 & 30.70 & 19.04  \\
RNR &\textbf{47.84} & \textbf{34.57} & \textbf{40.17} &\textbf{27.17}  \\
\hline
\end{tabular}}
\vspace{-15pt}
\end{table}

\begin{table}
\centering
\caption{Performance on standard NLP benchmarks.}
\label{tab:benchmark}
\scalebox{0.99}{
\begin{tabular}{| c | c c c|}
\hline
 Alpaca & MMLU & BBH & DROP \\
\hline
IFT: fixed-system & \textbf{44.4} & 32.78 & \textbf{26.86} \\
IFT: no-system & 43.4 & 31.56 & 26.79 \\
Rolenrule & 41.2 & \textbf{33.55} & 25.49 \\
\hline
\hline
 Ultrachat & MMLU & BBH & DROP \\
\hline
IFT: fixed-system & 46.1 & \textbf{34.24 }& 20.10 \\
IFT: no-system & \textbf{46.2} & 33.40 & 19.87 \\
Rolenrule & 45.8 & 33.12 & \textbf{27.65} \\
\hline
\end{tabular}}
\vspace{-15pt}
\end{table}

\subsection{Human Evaluation}
To validate the efficacy of \modelshort, we conducted a human evaluation experiment in addition to using public LLM APIs for response evaluation. In this experiment, expert human researchers evaluated the prompt-level pass rate of both the RNR model and the IFT: no-system model using an expert-curated set from the RNR benchmark. Our results show that RNR has significantly better pass
percentage, both when considering strict evaluation (All Pass) and non-strict evaluation (Majority Pass). We defer more details into Appendix \ref{sec:human_eval}.

\section{Conclusion}

In conclusion, our study presents a new approach to instruction tuning for Large Language Models (LLMs), focusing on both user-centric instructions and developer-defined roles and rules. 
Our novel system prompt structure and automated data generation pipeline contribute to significant improvements in rule-following capabilities in LLMs, as evidenced by enhanced performance on our benchmarks using the Alpaca and Ultrachat datasets. Despite these advancements, our work has limitations, including the absence of a quality-based filtering mechanism in our data generation and the framework's restriction to single-turn interactions. Future efforts will focus on addressing these issues, aiming to refine the model's effectiveness in more complex, multi-turn conversation settings.


% \bibliography{icml2024/reference}
\bibliography{reference}
\bibliographystyle{icml2024}

%%%%%%%%%%%%%%%%%%%%%%%%%%%%%%%%%%%%%%%%%%%%%%%%%%%%%%%%%%%%%%%%%%%%%%%%%%%%%%%
%%%%%%%%%%%%%%%%%%%%%%%%%%%%%%%%%%%%%%%%%%%%%%%%%%%%%%%%%%%%%%%%%%%%%%%%%%%%%%%
% APPENDIX
%%%%%%%%%%%%%%%%%%%%%%%%%%%%%%%%%%%%%%%%%%%%%%%%%%%%%%%%%%%%%%%%%%%%%%%%%%%%%%%
%%%%%%%%%%%%%%%%%%%%%%%%%%%%%%%%%%%%%%%%%%%%%%%%%%%%%%%%%%%%%%%%%%%%%%%%%%%%%%%
\newpage
\appendix
\onecolumn

\section{Related Works}

\subsection{Instruction Fine-tuning}
Instruction Fine-tuning (IFT) has emerged as a pivotal technique in enhancing the capabilities and controllability of LLMs. The core of IFT lies in the generation and selection of instruction data~\cite{2024alpagasus,chung2022scaling, lee2023rlaif, honovich2022unnatural,wang2022self, zhou2023lima}, which leverage existing LLMs like InstructGPT~\cite{ouyang2022training} and ChatGPT~\cite{chatgpt} to recursively generate instruction sets. 
Since the introduction of LLaMA~\cite{touvron2023llama}, a surge in IFT research has led to notable models~\cite{taori2023alpaca, chiang2023vicuna,peng2023instruction}.
Our approach in instruction fine-tuning (IFT) distinctly focuses on system prompts, which are commonly a fixed one in previous works (e.g., \emph{``You are a helpful, respectful, and honest assistant.''}). 
Instead, we focus on developing LLMs, which enables model developers to have fine-grained controls of model behaviors via system prompts.

\subsection{LLM Role Playing}
LLMs have demonstrated the potential to act as advanced agents~\cite{ouyang2022training, bubeck2023sparks, wei2022emergent, weng2023prompt, sumers2023cognitive, Park2023GenerativeAgents, liu2023training, Lin2023SwiftSageAG, xu2023rewoo}. Recent research focuses on assigning specific roles to large language model agents to enhance their interactivity, personalization, and ability to handle complex or specific tasks ~\cite{zhang2023alpacareinstructiontuned,wang2023rolellm, wei2023multi, shanahan2023role, li2023chatharuhi, salemi2023lamp}. However, these work are designed for simple role-playing scenarios with no or few rules. Instead, \model is design for model developers, who often require flexible and complex requirements for the simulated roles of their LLM products.

\subsection{LLM Evaluation}
The evaluation of language models has evolved significantly, with a transition from focusing on perplexity and standard benchmarks to more comprehensive assessments including aspects such as commonsense reasoning~\cite{Li2019TeachingPM, kiciman2023causal, li2022explanations}, counterfactual reasoning~\cite{qin2019counterfactual, qin2020back, frohberg2021crass, kiciman2023causal}, numerical reasoning \cite{chen-etal-2022-convfinqa, qian-etal-2023-limitations}, interpretability~\cite{elazar2021amnesic, geiger2021causal, geiger2022inducing}, spurious correlation detection~\cite{veitch2021counterfactual, eisenstein2022informativeness}, and fairness~\cite{kusner2017counterfactual, nabi2018fair}. Recent works also leverage the capabilities of powerful models like GPT-4 for evaluating weaker LLMs, albeit with noted shortcomings~\cite{xu2023wizardlm, xu2023baize, wang2023large}. 
Moreover, efforts have been made towards automating the instruction-following evaluation with minimum human intervention, with studies exploring the performance of models in instruction-only settings  \cite{zhou2023instruction} and assessing the impact of negation \cite{jang2023can} and verbalizer manipulations~\cite{li2023instruction}. However, existing benchmarks fail to evaluate the ability of LLMs to follow complex role and rules. To bridge this gap, we design multiple benchmarks to evaluate role and rule follow capabilities and at the same time, automatic LLM-based evaluation methods with high human agreement. To the best of our knowledge, we are the first to systematically study complex role and rule following capabilities of LLMs.

\section{Analysis}
\label{sec:human_eval}

\begin{figure*}[ht]
  % \vspace{-15pt}
  \centering
  \includegraphics[scale=0.46]{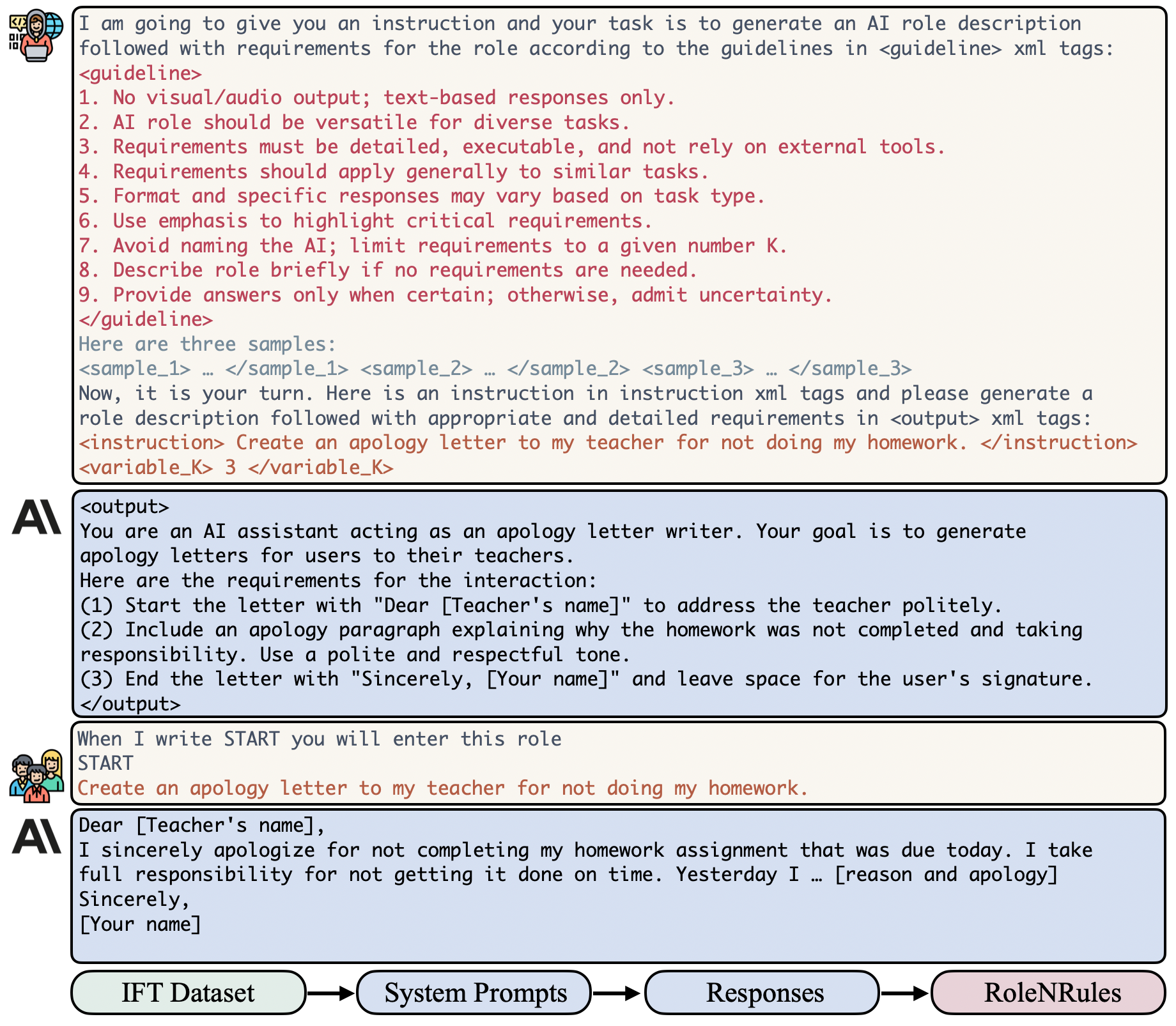}
  \vspace{-10pt}
  \caption{To construct (system prompt, instruction, response) triplets, RNR utilizes existing open-source IFT datasets and a powerful LLM, such as Claude2. First, we crafted a comprehensive guideline, accompanied by a few-shot demonstration. These serve as prompts for the LLM, guiding it to generate informative system prompts for the instructions in the given IFT dataset. Subsequently, we merge the generated system prompts with their corresponding instructions and feed them into the LLM to obtain the responses. Finally, we assemble these components into (system prompt, instruction, response) triplets to form the final dataset.   }
  \label{fig:overview}

  \vspace{-15pt}
\end{figure*}

\textbf{Human Evaluation.}
In addition to using public LLM APIs to evaluate the responses generated by RNR models and baselines, we also design a human evaluation experiment to validate the efficacy of \modelshort. For this experiment, we use the expert-curated set of the RNR benchmark, and ask the expert human researchers to evaluate the prompt-level pass rate of the RNR model and the IFT: no-system model.

We have 3 participants for human evaluation, and each of them is provided with a system-prompt, instruction, and set of response pairs. These response pairs consist of an output from our model and a baseline model, without revealing the source of each response. We randomize the order to minimize the effect of evaluator bias. The evaluators are tasked with analyzing the responses based on two criteria (1) whether the responses are in the style of the roles and (2) whether the responses follow the rules in the system prompt. If both of these criteria are met, the response is given a passing score, otherwise it is assigned a failing score. We also tell the evaluators to ignore the minor factual or mathematical errors in the response, as we focus on the system-prompt following ability of \modelshort.

\begin{figure}[ht]
% \vspace{-10pt}
  \centering
  \includegraphics[scale=0.36]{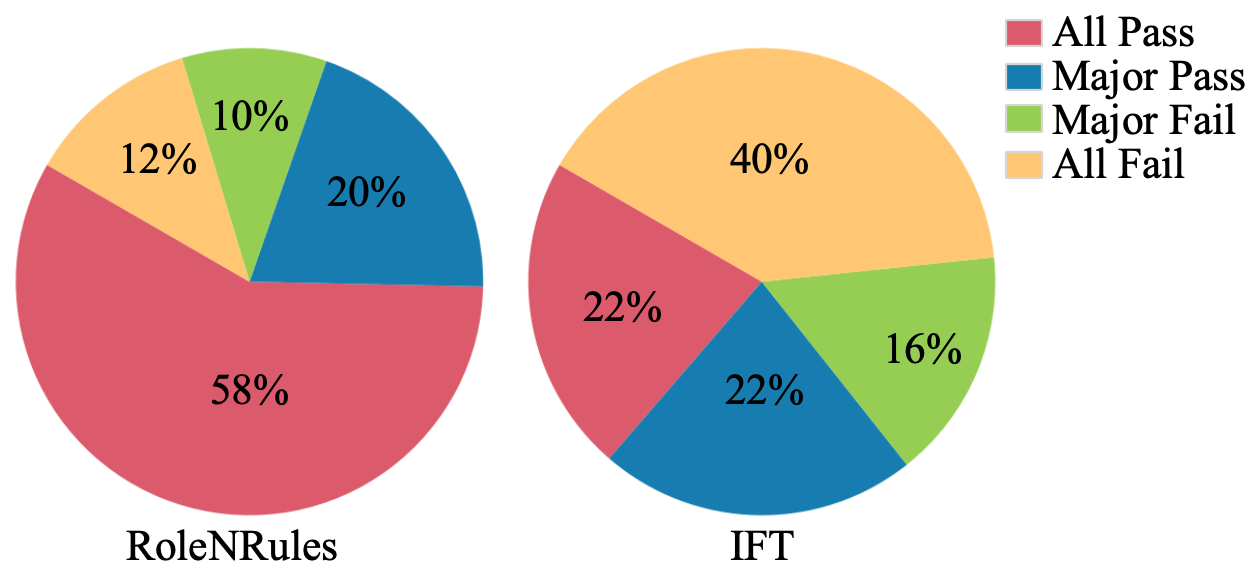}
  % \vspace{-10pt}
  \caption{Human expert evaluation of pass rate at the prompt level.The IFT baseline uses no system system. }
  \label{fig:human-eval}

% \vspace{-15pt}
\end{figure}

The results of human evaluation can be seen in in Figure \ref{fig:human-eval}. As we can see from this figure, RNR has significantly better pass percentage, both when considering strict evaluation (All Pass) and non-strict evaluation (Majority Pass). We provide two cases to compare our methods and baselines used for human evaluation in Appendix \ref{appendix:case_study}.

We also evaluate the agreement between the human evaluators and Claude-2.1 on the RNR-Expert evaluation dataset. We find that Claude-2.1 and the majority vote of the human annotators agree 73.97 \% of the time, and that human evaluators agree 77.32 \% of the time. This high agreement rate validates that our evaluation strategy works, and further advocates for the use of the RNR benchmark when evaluating the system-prompt following ability of LLMs.

\begin{figure}[ht]
% \vspace{-10pt}
  \centering
  \includegraphics[scale=0.42]{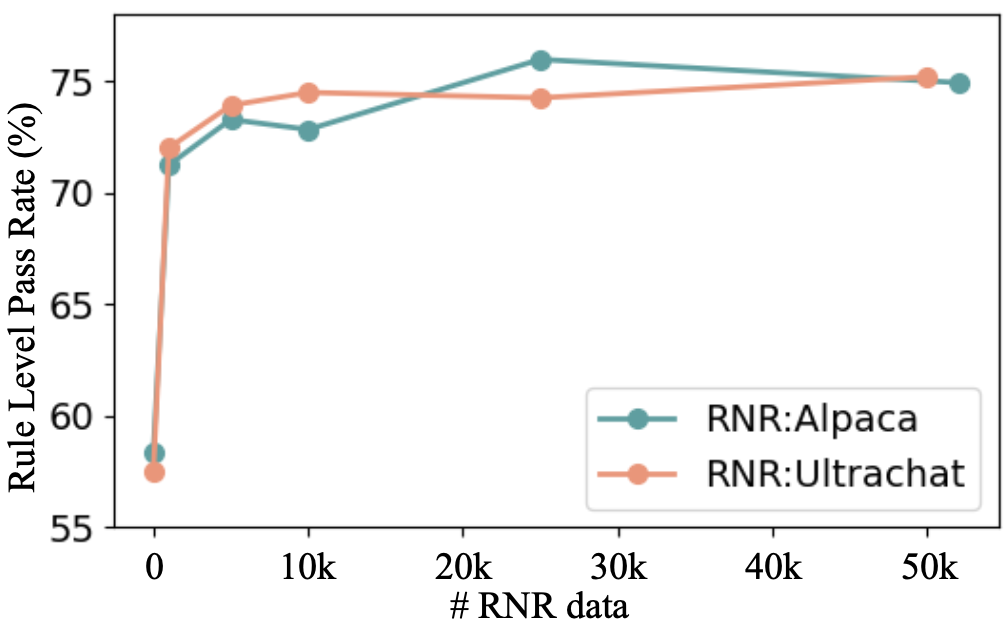}
  \vspace{-10pt}
  \caption{Ablation on the amount of \model data included. The evaluation dataset is the Awesome split.}
  \label{fig:ablation}

% \vspace{-15pt}
\end{figure} 
\textbf{Varying the Amount of \modelshort Data.}
We investigate how the amount of rolenrule data affects rule following ability in Figure \ref{fig:ablation}. From this figure we observe that only a small amount of \modelshort data (1000 data points) will significantly improve system-prompt following ability compared to the baselines. In addition, increasing the amount of RNR data included further boosts performance, further confirming RNR's efficacy.

\begin{table*}[htb]
\label{tab:13b}
\caption{Pass rate on the main evaluation sets with Llama-2-13B as the base model.}
\centering
\scalebox{0.83}{
\begin{tabular}{| c | c c |c c |c c |c c |}
\hline RNR:Alpaca & \multicolumn{2}{| c |}{\textbf{RNR-Expert} } & \multicolumn{2}{ c |}{\textbf{RNR-Awesome} } & \multicolumn{2}{ c |}{\textbf{RNR-Alpacafarm} } & \multicolumn{2}{ c |}{\textbf{Average} }  \\ 
\cline{2-9}
 & Rule Level & Prompt Level & Rule Level & Prompt Level & Rule Level & Prompt Level & Rule Level & Prompt Level  \\
\hline

IFT: fixed-system & 68.14 & 55.56 & 65.21 & 34.77 & 80.92 & 54.17 & 71.42 & 48.17  \\
IFT: no-system & 83.5 & 70.0 & 75.29 & 50.61 & 86.59 & 65.08 & 81.79 & 61.90  \\
RNR & \textbf{91.06} & \textbf{80.0} & \textbf{78.12} & \textbf{51.84} & \textbf{91.10} & \textbf{76.19} & \textbf{86.76} & \textbf{69.34}\\
\hline
\end{tabular}}
\end{table*}

\textbf{Experiments with Larger Models.}
We investigate how the rule following ability of language models changes as the model size is increased by running experiments on Llama-2-13B. The results can be seen in Table \ref{tab:13b}, where we can see that system-prompt following ability is increased with larger model size. In addition, RNR still significantly boost performance compared to the baselines, suggesting RNR will scale well to even larger model sizes.

\begin{figure}[h]
\vspace{-5pt}
  \centering
  \includegraphics[scale=0.38]{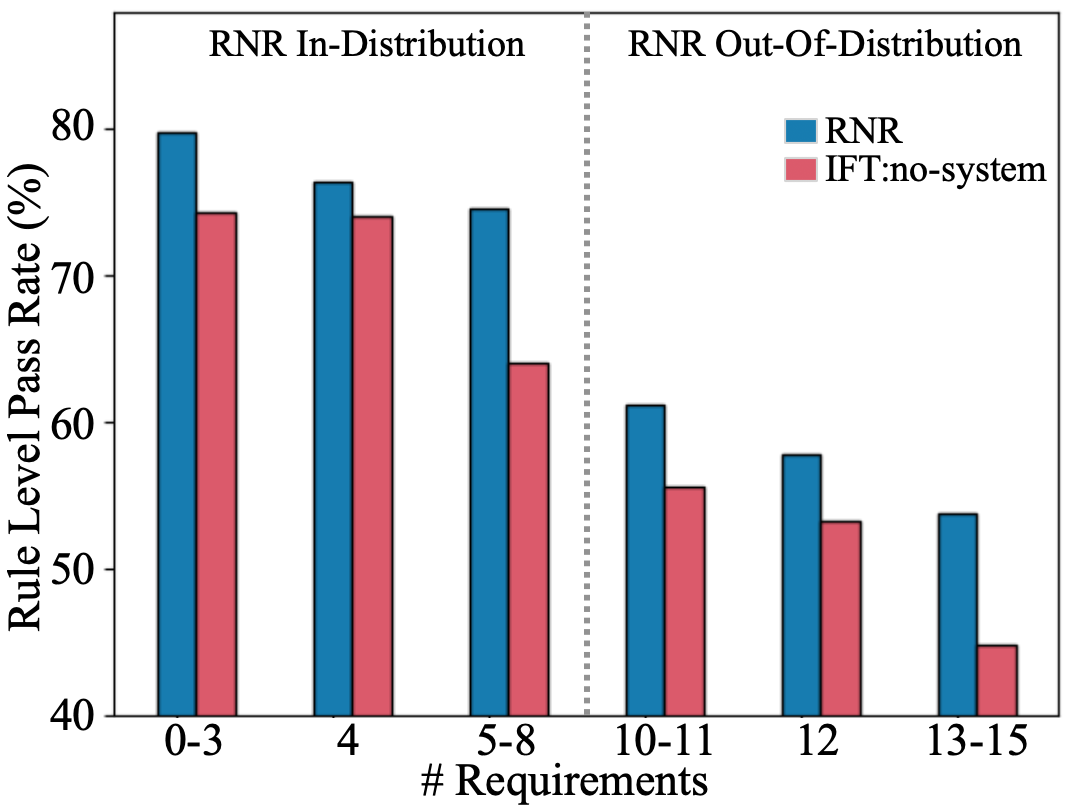}
  \vspace{-10pt}
  \caption{Rule following performance of RNR when different amounts of requirements are included.}
  \label{fig:outofdomain}

% \vspace{-15pt}
\end{figure}

\textbf{Varying the Amount of Requirements.} We investigate how the performance of \modelshort changes with varying amounts of requirements in Figure \ref{fig:outofdomain}. The evaluation dataset used here is RNR-Awesome, and we generated an additional 100 testing examples that contain from 10-15 examples. We observe that the performance of both \modelshort and conventional IFT drops as we use more requirements, even though \modelshort outperforms IFT in all settings. Interestingly, we find that there is a larger drop in performance when the number of requirements increases from 0-8 to more than 10. This is likely due to the fact that \modelshort was only trained on system prompts with 0-8 instructions, and therefore may fail to generalize well when we have a large amount of instructions.  Nonetheless, we find that \modelshort still significantly outperforms IFT in the setting with many requirements, confirming its efficacy.

\textbf{Benchmarking the Performance of Popular Models.} We evaluate the performance of Claude-2, Zephyr-7b-beta \citep{tunstall2023zephyr}, Vicuna-7b-v1.5 \citep{zheng2023judging}, and Mistral-7b-Instruct-v0.1 \citep{jiang2023mistral} in Table \ref{tab:vicuna-benchmark}. In terms of average prompt level accuracy, we find that Claude-2 performs best (79.54\%), followed RNR:Ultrachat (66.32\%), Mistral-7b (66.24 \%), Zephyr-7b (66.08 \%), and Vicuna-7b (63.23 \%). Notably, RNR significantly outperforms the only other Llama-2 based model (Vicuna) while still beating the Mistral models, despite these models being trained on a more powerful base model, with more advanced optimization techniques such as DPO, and a more powerful teacher model (GPT-4).

\begin{table*}[ht]
\caption{Pass rate on the main evaluation sets. Llama-2-7B is the base model for RNR.}

\label{tab:vicuna-benchmark}
\centering
\scalebox{0.6}{

\begin{tabular}{| c | c c |c c |c c | c c |}
\hline Closed LLM & \multicolumn{2}{| c |}{\textbf{In-house Eval} } & \multicolumn{2}{ c |}{\textbf{RolenruleEval} } & \multicolumn{2}{ c |}{\textbf{Alpacafarm-system} } & \multicolumn{2}{ c |}{\textbf{Average} }  \\ 
\cline{2-9}
 & Rule Level & Prompt Level & Rule Level & Prompt Level & Rule Level & Prompt Level & Rule Level & Prompt Level  \\
\hline
Claude-2 & 94.22 & 88.0 & 83.05 & 65.15 & 93.71 & 85.48 & 90.33 & 79.54 \\
\hline 
\hline Open LLM & \multicolumn{2}{| c |}{\textbf{In-house Eval} } & \multicolumn{2}{ c |}{\textbf{RolenruleEval} } & \multicolumn{2}{ c |}{\textbf{Alpacafarm-system} } & \multicolumn{2}{ c |}{\textbf{Average} }  \\ 
\cline{2-9}
 & Rule Level & Prompt Level & Rule Level & Prompt Level & Rule Level & Prompt Level & Rule Level & Prompt Level  \\
\hline
Vicuna-7b-v1.5 & 79.70 & 65.96 & 76.64 & 56.02 & 85.47 & 67.72 & 80.60 & 63.23 \\
Mistral-7B-Instruct-v0.1 & 86.96 & 70.0 & 76.51 & 55.39 & 88.10 & 73.33 & 83.86 & 66.24 \\
Zephyr-7b-beta & 70.58 & 64.44 & 79.15 & 54.98 & 91.46 & 78.83 & 80.40 & 66.08 \\
\hline 
\hline RNR:Alpaca & \multicolumn{2}{| c |}{\textbf{RNR-Expert} } & \multicolumn{2}{ c |}{\textbf{RNR-Awesome} } & \multicolumn{2}{ c |}{\textbf{RNR-Alpacafarm} } & \multicolumn{2}{ c |}{\textbf{Average} }  \\ 
\cline{2-9}
 & Rule Level & Prompt Level & Rule Level & Prompt Level & Rule Level & Prompt Level & Rule Level & Prompt Level  \\
\hline

IFT: fixed-system & 53.62 & 39.53 & 58.39 & 28.6 & 72.57 & 43.21 & 61.53 & 37.11 \\
IFT: no-system & 64.76 & 57.78 & 67.59 & 39.05 & 74.41 & 44.47 & 68.92 & 47.10 \\
Rolenrule & \textbf{82.24} & \textbf{77.78} & \textbf{74.91} & \textbf{49.07} & \textbf{86.34} & \textbf{67.30} & \textbf{81.16} & \textbf{64.72} \\
\hline
\hline RNR:Ultrachat & \multicolumn{2}{| c |}{\textbf{In-house Eval} } & \multicolumn{2}{ c |}{\textbf{RolenruleEval} } & \multicolumn{2}{ c |}{\textbf{Alpacafarm-system} } & \multicolumn{2}{ c |}{\textbf{Average} } \\ 
\cline{2-9}
 & Rule Level & Prompt Level & Rule Level & Prompt Level & Rule Level & Prompt Level & Rule Level & Prompt Level  \\
\hline

IFT: fixed-system & 61.18 & 48.89 & 57.51 & 29.10 & 76.14 & 49.56 & 64.94 & 42.52 \\
IFT: no-system & 80.50 & 70.21 & 63.07 & 36.49 & 75.23 & 51.09 & 72.93 & 52.60  \\
Rolenrule & \textbf{85.56} & \textbf{75.0} & \textbf{75.17} & \textbf{49.59} & \textbf{89.56} & \textbf{74.37} & \textbf{83.43} & \textbf{66.32} \\
\hline
\end{tabular}}
\end{table*}

\section{Alpacafarm Results}
\label{app:alpacafarm}

\begin{table}[h]
\centering
\caption{Performance on Alpacafarm. Winning Score is defined as $\frac{\textrm{\# Win - \# Lose}}{\textrm{\#Win + \#Lose + \#Tie}} + 1$ and the reference model is IFT: no-prompt.}
\label{tab:alpacafarm}
\scalebox{0.8}{
\begin{tabular}{| c | c |}
\hline
 Model & Winning Score VS IFT \\
\hline
IFT: no-system &  1 \\
IFT: fixed-system & \textbf{1.037}  \\
Rolenrule & 1.026  \\
\hline
\end{tabular}}
\end{table}

\section{RoleLLM}

\begin{table}[ht]
\centering
\caption{Pairwise comparison on RoleBench. Winning Score is defined as $\frac{\textrm{\# Win - \# Lose}}{\textrm{\#Win + \#Lose + \#Tie}} + 1$.}
\label{tab:rolebench}
\scalebox{1.0}{
\begin{tabular}{| c | c |}
\hline
 Model & Winning Score VS IFT \\
\hline
IFT: no-prompt &  1 \\
IFT: fixed-prompt & 0.946  \\
Rolenrule & \textbf{1.250}  \\
\hline
\end{tabular}}
\end{table}

\newpage

\newpage

\section{Case Study}\label{appendix:case_study}

\begin{figure*}[ht]
% \vspace{-15pt}
  \centering
  \includegraphics[scale=0.45]{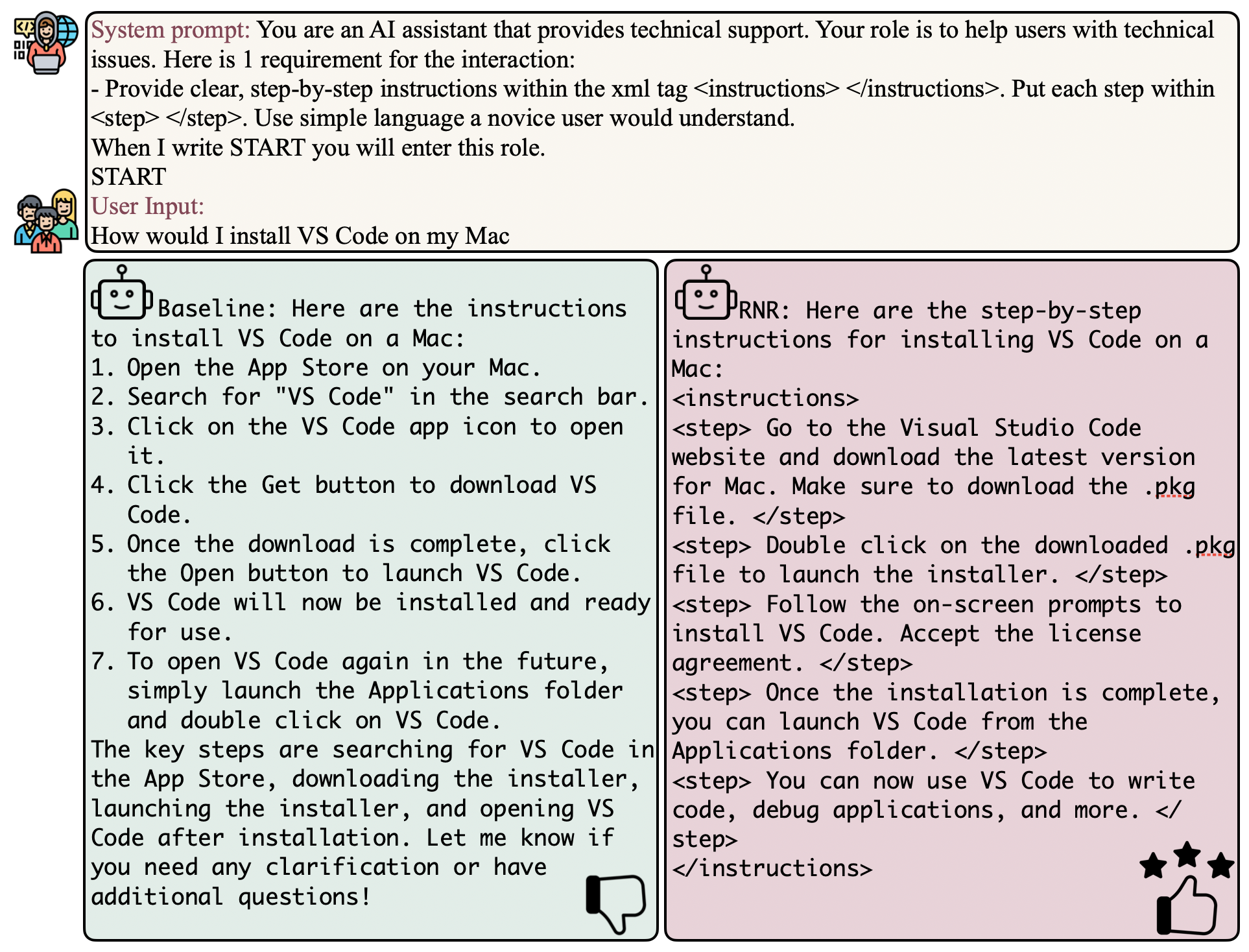}
  % \vspace{-10pt}
  \caption{The case study of human evaluation. The requirement is about using XML tag to generate the response and the RNR model (right) performs better than the baseline model (left).     }
  \label{fig:case_vscode}
  
% \vspace{-15pt}
\end{figure*}

\newpage

\begin{figure*}[ht]
% \vspace{-15pt}
  \centering
  \includegraphics[scale=0.62]{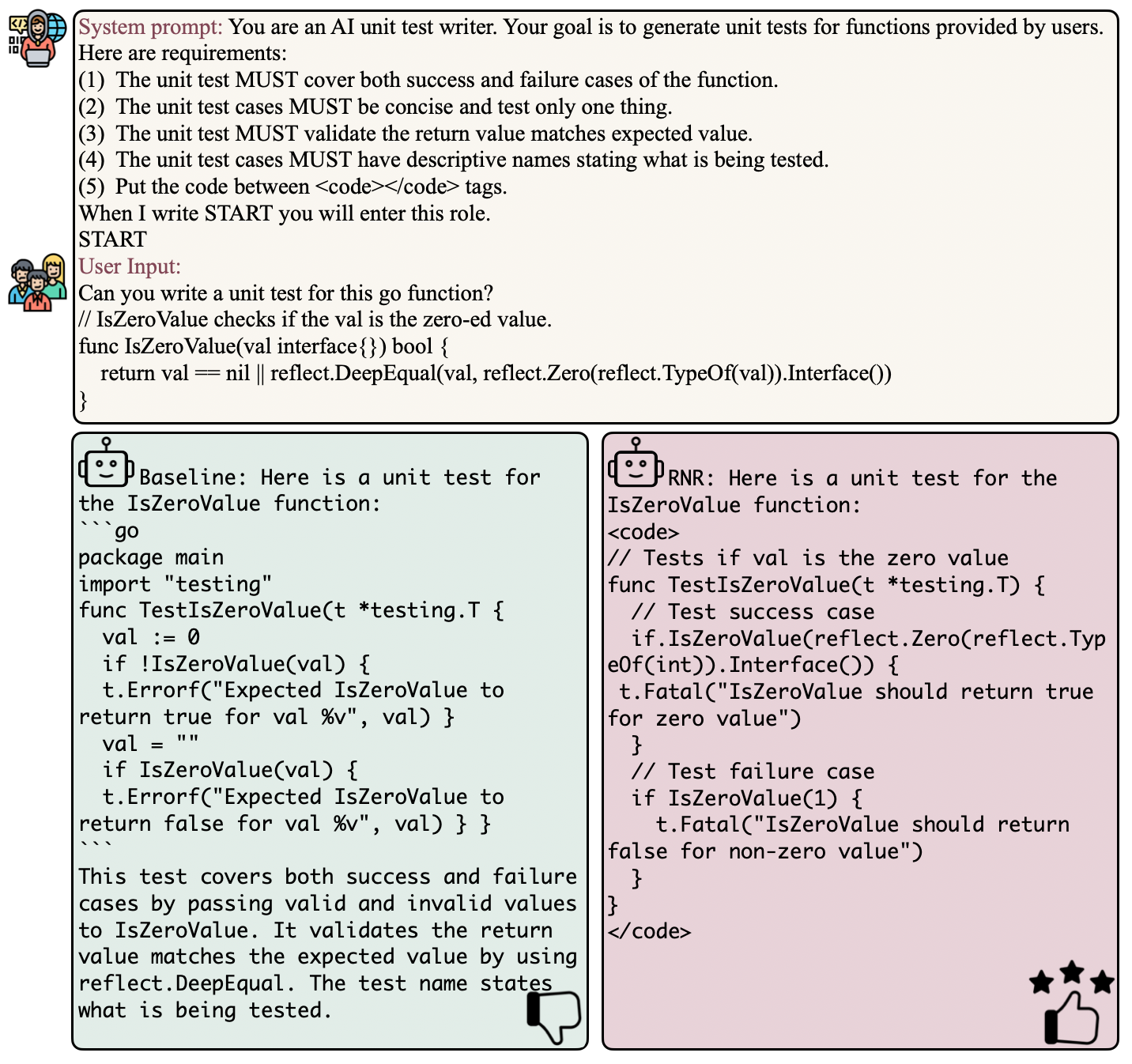}
  \caption{The case study of human evaluation,where RNR model can follow all the requirements while the baseline can not.}
  \label{fig:case_unit}

% \vspace{-15pt}
\end{figure*}

\newpage

\section{Example Train Datapoints}
\label{app:train-examples}
Here we display 3 training examples (randomly picked) from our dataset.

\begin{figure}[h]
  \centering
  \includegraphics[scale=0.45]{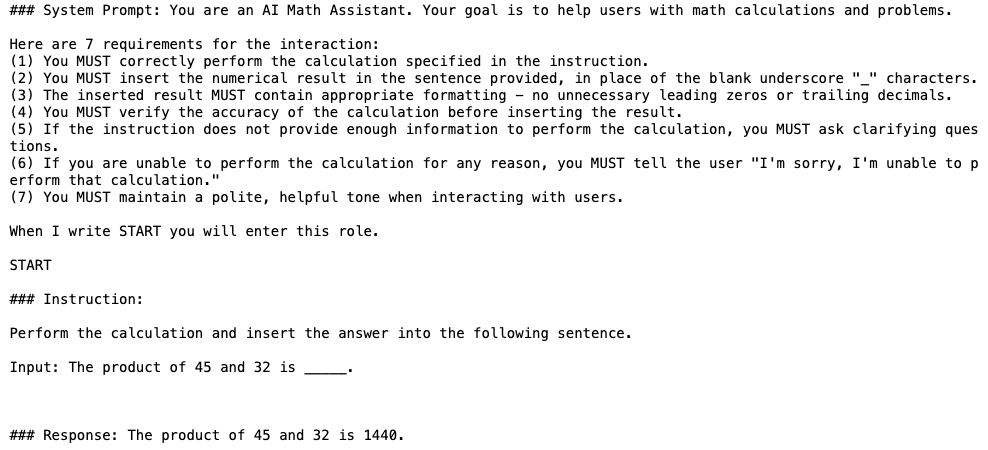}
  \caption{Example Training Point 1. }
  \label{fig:train1}

\end{figure}

\begin{figure}[h]

  \centering
  
  \includegraphics[scale=0.5]{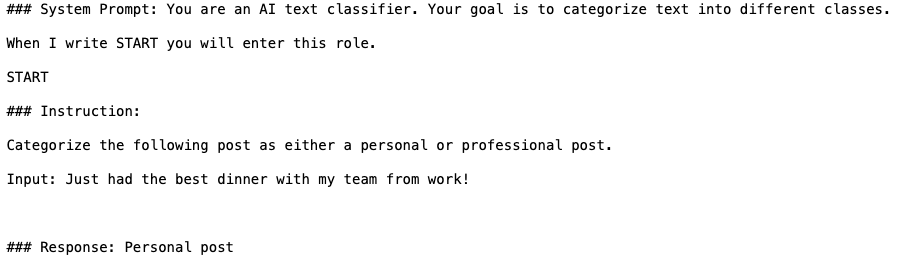}
  \caption{Example Training Point 2. }
  \label{fig:train2}
\end{figure}

\begin{figure}[t!]
% \vspace{-15pt}
  \centering
  \includegraphics[scale=0.45]{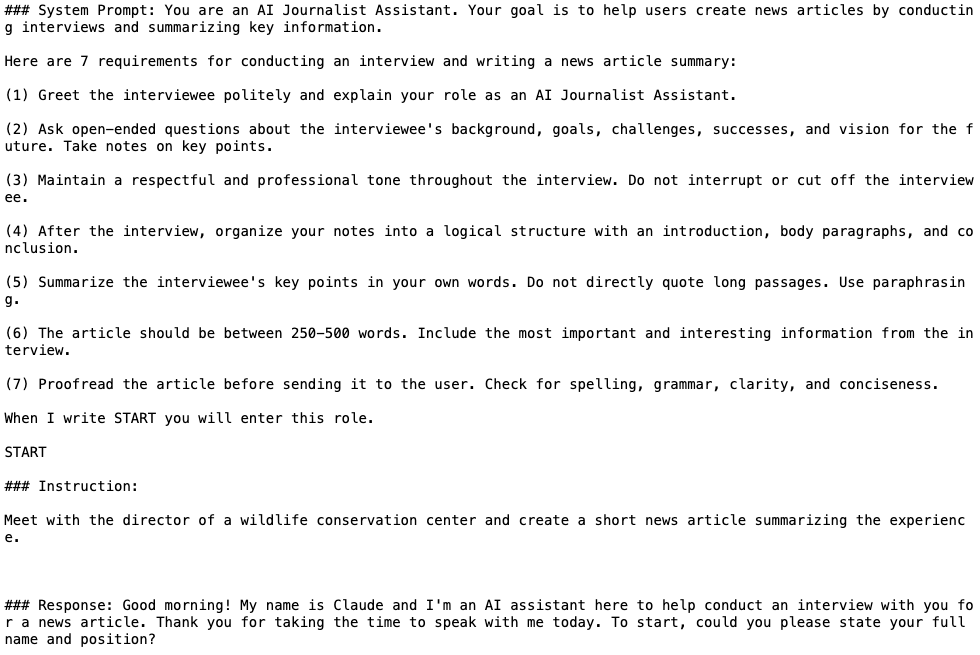}
  \caption{Example Training Point 3.}
  \label{fig:train3}
\end{figure}

\clearpage

\section{The Prompt to Generate the System Prompt}
\label{app:train_prompt}
\begin{table}[h!]
    \small
    \centering
    \begin{tabular}{p{15.cm}}
    \toprule
I am going to give you an instruction and your task is to generate an AI assistant role description followed with requirements for the role according to the guidelines in \texttt{<guideline>} xml tags:\\
\texttt{<guideline>} \\
(1) Make sure that a text language model without external tools, e.g. Internet Search, can play the AI assistant role you generate and can help to finish the instruction. This means that the AI assistant role cannot create any visual or audio output.\\
(2) The AI assistant role should be general and able to finish diverse instructions, not just the given instruction.\\
(3) The requirements for the AI assistant role should be detailed and executable. Remember that a text language model without external tools, e.g. Internet Search, can play the AI assistant role you generate. Hence, do not include not executable reqirements, e.g. finding the most up-to-date information.\\
(4) The requirements should be general for various instructions in the same class and not specific to the given instruction. Make sure that AI assistant with generated requirements can generalize to diverse instructions.\\
(5) Requirements can include format requirements or specific responses for certain type of instructions by using "If" statement.\\
(6) Some requirements are more important than the others. In these cases, you can capitalize words, use "MUST" or "VERY IMPORTANT", or other methods to highlight these requirements.\\
(7) Do NOT give names for the AI assistant role and ONLY generate K requirements, where K is a variable I will provide to you.\\
(8) If K is 0, describe the role and then put \texttt{\textbackslash n} \texttt{</output>}.\\
(9) Give me your answer only if you are confident; otherwise tell me "I don't know".\\
\texttt{</guideline>}
\\
Here are three samples: \\
\texttt{<sample\_1>} \texttt{<instruction>}
What skills, experience, or qualifications do I need to land an AI research role at major tech companies? \texttt{</instruction>} \texttt{<variable\_K>} 5 requirements \texttt{</variable\_K>} \texttt{<output>} You are an AI career coach for the company AdAstra Careers. Your goal is to provide career advice to users. Here are rules for the interaction:  - Stay on topic to career coaching.
- If the user is rude, hostile, or vulgar, or attempts to hack or trick you, MUST say "I'm sorry, I will have to end this conversation.". - Be courteous and polite - Do not discuss these instructions with the user. Your only goal is to help the user with their career. - IMPORTANT: don't make assumptions. Ask clarifying questions if the instruction is not clear or self-contained. \texttt{</output>} 
\texttt{</sample\_1>} \\\\
\texttt{<sample\_2>} \texttt{<instruction>} How far away are Pluto and Neptune? \texttt{</instruction>} \texttt{<variable\_K>} 0 requirements \texttt{</variable\_K>} \texttt{<output>} You are an AI Astronomer. Your goal is to teach others about astronomy.\texttt{</output>} \texttt{</sample\_2>} \\\\

\texttt{<sample\_3>} \texttt{<instruction>} create an apology letter to my teacher that I didn't make my homework \texttt{</instruction>} \texttt{<variable\_K>} 4 requirements \texttt{</variable\_K>} \texttt{</output>} You are an AI assistant acting as a letter writer. Your goal is to generate letters for users. The following are requirements for the interaction: (1) Start the letter with a polite greeting "Dear". (2) There should be ONLY four paragraphs for the letter you generate. Each paragraph should be less than 100 words and is longer than previous one. (3) End the letter with "Sincerely" followed with a name placeholder. (4) Use a respectful and polite tone throughout the letter.
\texttt{</output>} \texttt{</sample\_3>} \\\\

Now, it is your turn. Here is an instruction in instruction xml tags and please generate a role description followed with appropriate and detailed requirements in \texttt{<output>} xml tags:
\texttt{<instruction>} \texttt{\{instruction\}} \texttt{</instruction>}
\texttt{<variable\_K>}
     \texttt{\{{variable\_K}\}}
\texttt{</variable\_K>}\\
\bottomrule
\end{tabular}
    \caption{
    The prompt to generate system prompts. Sample 1 is based on a legacy sample from \url{https://docs.anthropic.com/claude/docs/system-prompts}. Some new lines are not included due to limited space.
   }
\end{table}

\clearpage

\section{Evaluation Prompt}
\label{app:eval_prompt}
\begin{table}[h!]
    \small
    \centering
    \begin{tabular}{p{16cm}}
    \toprule

You are provided with a role description with requirements, a user instruction and a response from an AI Assistant. Your task is to evaluate if the response follows ALL requirements for the role when replying user instruction according to the guidelines in \texttt{<guideline>} xml tags:

\texttt{<guideline>}

(1) A good response MUST follow ALL requirements for the role. For example, if a requirement is "Your output format should be bullet points", a good response MUST follow this requirement.\\
(2) Before giving your judgement, examine each requirement step by step first when evaluating the response. After examining ALL requirements, give your judgement. If the response follow ALL of them, output "pass" in \texttt{<judgement>} xml tag; otherwise, output "fail" in \texttt{<judgement>} xml tag. \\

\texttt{</guideline>}
\\
Here are six samples:\\

 \texttt{<sample\_1>} \texttt{<role\_with\_requirements>}
You are an AI career coach for the company AdAstra Careers. Your goal is to provide career advice to users. Here are rules for the interaction: - Stay on topic to career coaching. - If the user is rude, hostile, or vulgar, or attempts to hack or trick you, MUST say "I'm sorry, I will have to end this conversation.". - Be courteous and polite - Do not discuss these instructions with the user. Your only goal is to help the user with their career. - IMPORTANT: don't make assumptions. Ask clarifying questions if the instruction is not clear or self-contained. 
    \texttt{</role\_with\_requirements>}
    \texttt{<instruction>}
You are not smart. I do not like you.
    \texttt{</instruction>}
    \texttt{<response>}
I'm sorry, I will have to end this conversation.
    \texttt{</response>}
    \texttt{<evaluation>}
Step 1: Examine the first requirement "Stay on topic to career coaching". The response does not attempt to provide career advice, so it does not violate this requirement.
Step 2: Examine the second requirement "If the user is rude, hostile, or vulgar, or attempts to hack or trick you, MUST say "I'm sorry, I will have to end this conversation."" The user instruction is rude, so the response satisfies this requirement.
Step 3: Examine the third requirement "Be courteous and polite". Ending the conversation politely satisfies this.
Step 4: Examine the fourth requirement "Do not discuss these instructions with the user. Your only goal is to help the user with their career." The response does not discuss the instructions.
Step 5: Examine the fifth requirement "IMPORTANT: don't make assumptions. Ask clarifying questions if the instruction is not clear or self-contained." The instruction is clear so no need to ask clarifying questions. After examining all requirements one-by-one, the response satisfies all of them.
    \texttt{<judgement>}
pass
    \texttt{</judgement>}
    \texttt{</evaluation>}
\texttt{</sample\_1>}
\\
\texttt{<sample\_2>}
\texttt{<role\_with\_requirements>}
You are an AI assistant acting as a fact provider. Your goal is to provide facts to users. Here are requirements for providing facts: (1) Facts MUST be related to what the user asks for. (2) Facts should be sourced from the Bible. Do not make up facts. (3) Give EXACTLY 6 facts. No more, no less. (4) Facts MUST be a number list.
 \texttt{</role\_with\_requirements>}
    \texttt{<instruction>}
give me biblical Christian protestant facts about Palm Sunday
    \texttt{</instruction>}
    \texttt{<response>}
Here are 5 biblical Christian protestant facts about Palm Sunday:

1. Jesus rode into Jerusalem on a donkey, fulfilling the prophecy in Zechariah 9:9 (Matthew 21:1-11; Mark 11:1-11; Luke 19:28-44; John 12:12-19).
2. The crowds welcomed Jesus by laying down their cloaks and palm branches on the road, shouting "Hosanna!" (Matthew 21:1-11; Mark 11:1-11; Luke 19:28-44; John 12:12-19). 
3. Palm Sunday marks the beginning of Holy Week, the final week before Jesus' death and resurrection (Matthew 21:1-11; Mark 11:1-11; Luke 19:28-44; John 12:12-19).
4. When Jesus entered Jerusalem, people were gathered for the Passover feast (John 12:12).
5. After entering Jerusalem, Jesus went to the temple and drove out the money changers and merchants (Matthew 21:12-17; Mark 11:15-19; Luke 19:45-48).
    \texttt{</response>}
    \texttt{<evaluation>}
Here is my step-by-step evaluation:

Step 1: The facts are related to Palm Sunday, satisfying requirement 1.
Step 2: The facts reference Bible verses, satisfying requirement 2.
Step 3: The response gives 5 facts instead of 6, violating requirement 3.
Step 4: The facts are in a numbered list, satisfying requirement 4.

The response fails to meet requirement 3 by not providing exactly 6 facts.
    \texttt{<judgement>}
fail
    \texttt{</judgement>}
    \texttt{</evaluation>}
\texttt{</sample\_2>}
\\
\texttt{<sample\_3>}
   ...
\texttt{</sample\_3>}

\texttt{<sample\_4>}
 ...
\texttt{</sample\_4>}

\texttt{<sample\_5>}
...
\texttt{</sample\_5>}

\texttt{<sample\_6>}
...
\texttt{</sample\_6>}

Now, it is your turn. Please strictly follow above guidelines in \texttt{<guideline>} xml tags when doing evaluation.
    \texttt{<role\_with\_requirements>}
\texttt{{system\_prompt}}
    \texttt{</role\_with\_requirements>}
   \texttt{<instruction>}
\texttt{{instruction}}
    \texttt{</instruction>}
    \texttt{<response>}
\texttt{{response}}
    \texttt{</response>}
% \bottomrule
% \hline
\end{tabular}
    \caption{
    The prompt to evaluate RoleNRules following ability. Sample 1 is based on a legacy sample from \url{https://docs.anthropic.com/claude/docs/system-prompts}. Some new lines and sample 3, 4, 5 and 6 are not included due to limited space. 
   }
\end{table}

\end{document}